\documentclass[conference]{IEEEtran}
\IEEEoverridecommandlockouts
\usepackage{cite}
\usepackage{amsmath,amssymb, amsfonts}
\usepackage{graphicx}
\usepackage{textcomp}
\usepackage{xcolor}
\usepackage[nolist]{acronym}
\usepackage{balance}
\usepackage{adjustbox}
\usepackage{subfig}
\usepackage[breaklinks, hidelinks]{hyperref}
\usepackage[justification=centering]{caption}
\def\BibTeX{{\rm B\kern-.05em{\sc i\kern-.025em b}\kern-.08em
    T\kern-.1667em\lower.7ex\hbox{E}\kern-.125emX}}

\title{Sphere-GAN: a GAN-based approach for saliency estimation in 360\textdegree\ videos\\
\thanks{This work was supported by the European Union under the Italian National Recovery and Resilience Plan (NRRP) Mission 4, Component 2, Investment 1.3, CUP C93C22005250001, partnership on “Telecommunications of the Future” (PE00000001 - program “RESTART”)”.}
}
\author{\IEEEauthorblockN{Mahmoud Z. A. Wahba, Sara Baldoni, and Federica Battisti}
\IEEEauthorblockA{\textit{Department of Information Engineering} \\
\textit{University of Padova}\\
Padua, Italy \\
mahmoudza.wahba@phd.unipd.it, sara.baldoni@unipd.it, federica.battisti@unipd.it}
}
\begin{acronym}
\acro{BCE}{Binary Cross Entropy}
\acro{CC}{Correlation Coefficient}
\acro{CMP}{Cube Map Projection}
\acro{CNN}{Convolutional Neural Network}
\acro{DCNN}{Deep Convolutional Neural Network}
\acro{DNN}{Deep Neural Networks}
\acro{ERP}{Equi-Rectangular Projection}
\acro{FoV}{Field of View}
\acro{GAN}{Generative Adversarial Network}
\acro{MSE}{Mean Squared Error}
\acro{NSS}{Normalized Scanpath Saliency}
\acro{VR}{Virtual Reality}
\end{acronym}
\begin{document}
%
\maketitle
\begin{abstract}
The recent success of immersive applications is pushing the research community to define new approaches to process 360\textdegree\ images and videos and optimize their transmission. 
Among these, saliency estimation provides a powerful tool that can be used to identify visually relevant areas and, consequently, adapt processing algorithms. Although saliency estimation has been widely investigated for 2D content, very few algorithms have been proposed for 360\textdegree\ saliency estimation. Towards this goal, we introduce Sphere-GAN, a saliency detection model for 360\textdegree\ videos that leverages a Generative Adversarial Network with spherical convolutions. 
Extensive experiments were conducted using a public 360\textdegree\ video saliency dataset, and the results demonstrate that Sphere-GAN outperforms state-of-the-art models in accurately predicting saliency maps. 
\end{abstract}
\section{Introduction}
\label{sec:intro}

With the increasing diffusion of \ac{VR} systems and the cost reduction of omni-directional cameras, 360\textdegree\ content is becoming widespread.
Despite this, its processing, storage, and transmission still pose several challenges. 

Concerning processing, the standard techniques employed for 2D images and videos cannot be directly applied to the 360\textdegree\ content. Although the spherical image or frame can be mapped to a 2D equivalent through projections, such as \ac{ERP} or \ac{CMP}, this operation results in an increase in geometric distortions and in a consequent decrease in performance. This applies also to \acp{DNN}, which are defined for traditional euclidean data~\cite{Xu_PAMI_2022, Yang_TCSVT_2024}.
However, the design of processing methods that work in the spherical domain requires an adaptation of existing 2D algorithms.
As to storage and transmission, 360\textdegree\ content requires a larger amount of memory and a significantly higher bitrate. More specifically, the bandwidth required to stream a 360\textdegree\ video is an order of magnitude higher with respect to its 2D counterpart: while a traditional 4K video requires about 25 Mb/s, the data rate reaches 400 Mb/s for streaming a 4K 360\textdegree\ video to each eye~\cite{Zink_IEEE_2019}. Moreover, to achieve full immersion, the round-trip time should reach $10$~ms~\cite{ITU_743-10}.
The transmission burden can be reduced considering that, although the 360\textdegree\ video is by nature omni-directional, users will focus on one portion at a time, due to the limitations of the human visual system. This portion roughly corresponds to $20\%$ of the content, and is commonly referred to as viewport. Therefore, viewport prediction techniques can be exploited to reduce the need of transmitting the entire frame at any time instant~\cite{Yaqoob_CST_2020}. 
Viewport prediction can be performed by analyzing user head movements and applying a prediction algorithm~\cite{Wang_IoT_2024}. However, it has been demonstrated that saliency estimation is a key enabler for increasing the performance of head movement-based viewport prediction~\cite{Setayesh_ICME_2023}.

Saliency estimation is the analysis of the relevance of different portions of an image/frame according to a human observer~\cite{Battisti_SPIC_2018}. It represents a well-known issue for 2D images and videos and an active research field for 360\textdegree\ content. The available techniques can be classified in feature-based and learning-based approaches. Examples of the first kind can be found in~\cite{Fang_SPIC_2018, Baldoni_ISPA_2023}. As manual feature selection is challenging and sometimes ineffective, researchers shifted toward learning-based techniques. In this case, it is possible to distinguish between approaches based on standard convolutions, such as~\cite{Zou_ToB_2023, Cong_TNNLS_2024}, or methods that extend this concept to the spherical domain. 
\begin{figure*}[htb]
    \centering
    \includegraphics[width=\linewidth]{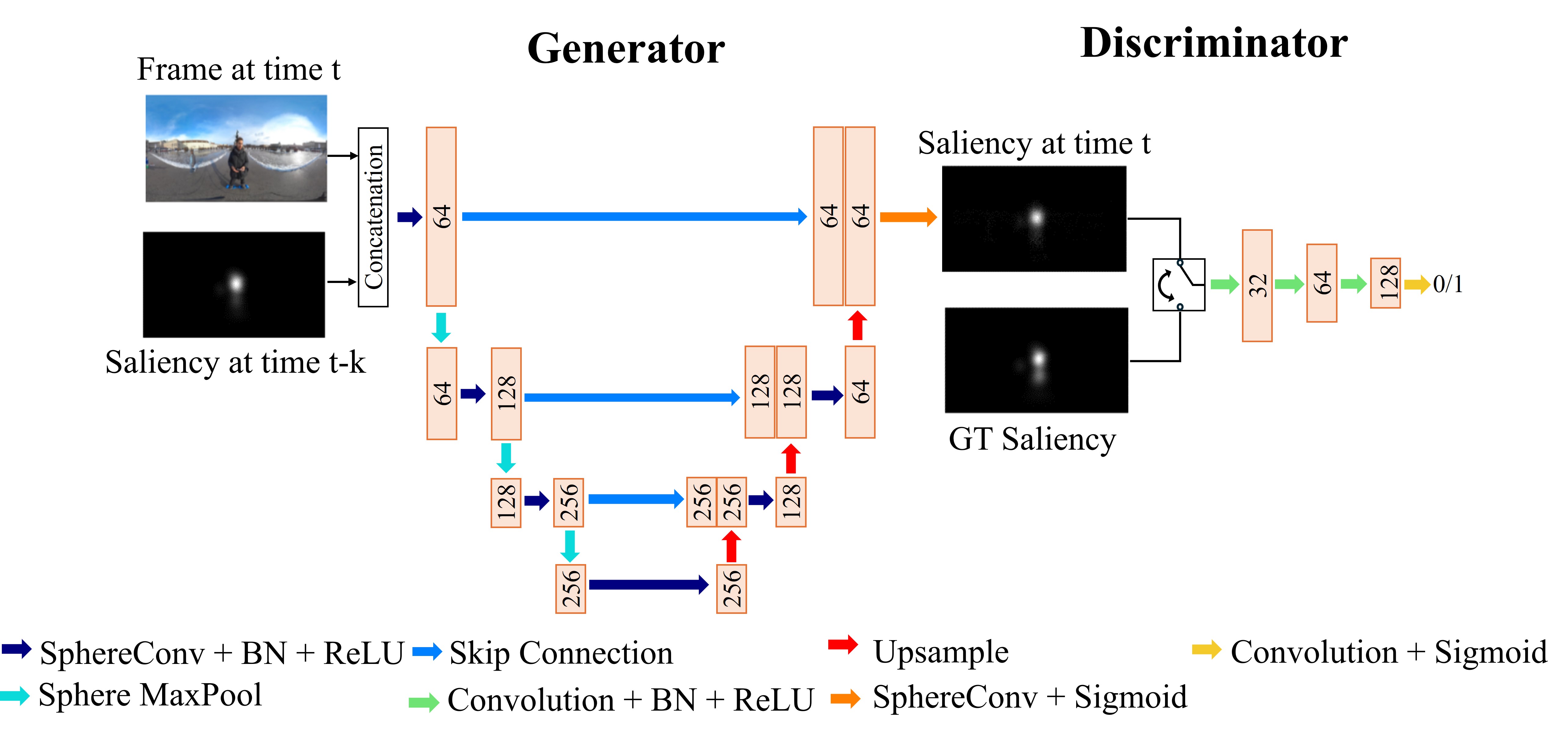}
    \caption{Sphere-GAN model structure.} 
    \label{fig:GAN}
\end{figure*}
In the latter direction, in~\cite{Coors_ECCV_2018, Zhang_ECCV_2018} two methods for spherical convolutions have been investigated. 
In the former, the authors present SphereNet, which transfers convolution and pooling operations from the 2D domain to the spherical one. This is done by representing the kernel as a small patch tangent to the sphere, thus avoiding discontinuities at the poles. The presented method has been tested for image classification and object detection tasks. In~\cite{Zhang_ECCV_2018}, spherical convolutions have been directly applied to saliency estimation by introducing the SphericalU-Net model. The kernel has been defined on a spherical crown, and the convolution has been realized by performing rotations of the kernel along the sphere, thus allowing parameter sharing. In addition, the authors introduced a dataset of $104$ 360\textdegree\ videos viewed by $20$ observers and achieved a correlation of $0.6246$ between the estimated and ground-truth saliency maps.
The same dataset was employed in~\cite{Peng_IWCMC_2023}, where the authors proposed a saliency model based on SphereNet. They introduced SphereU-Net, applying the principles of SphereNet to a U-Net, and achieved an improvement in correlation up to $0.8368$.
In this work, we introduce a \ac{GAN}-based approach for saliency estimation employing spherical convolutions. To this aim, we leverage the findings presented in~\cite{Coors_ECCV_2018, Zhang_ECCV_2018, Peng_IWCMC_2023} and expand the application of spherical convolutions to \acp{GAN}. To the best of our knowledge, this is the first time that spherical convolutions are employed in a \ac{GAN}-based architecture, and this represents a key contribution for video saliency estimation. 
We employ the Sphere-Unet architecture as generator and add an ad-hoc defined discriminator. The proposed approach shows increased performance with respect to the baseline methods on a publicly available dataset, thus demonstrating its effectiveness. 

\section{Proposed method}
\label{sec:pagestyle}
To store and transmit spherical images, a projection operation, such as the \ac{ERP}, is typically performed to convert the spherical image into a flat image. 
The conversion introduces significant distortions, particularly near the poles, making saliency estimation based on standard convolutions on equi-rectangular images ineffective.
For this reason, we present Sphere-GAN, which extends the application of the spherical convolutions proposed in~\cite{Coors_ECCV_2018} to \ac{GAN} architectures.

\subsection{Network Structure}
Our Sphere-GAN, uses the U-Net proposed in~\cite{Peng_IWCMC_2023}
as generator. 
U-Net features encoder and decoder paths with skip connections, allowing the decoder to access high-resolution features lost during the encoder's down-sampling process. The encoder comprises 4 spherical convolution layers, batch normalization layers, ReLU activation functions, and 3 spherical max-pooling layers. The spherical convolutions have a $3\times3$ window size with a stride of $1$. Each convolution layer, followed by batch normalization layer and ReLU activation function, is succeeded by a spherical max-pooling layer with a $2\times2$ window size and a stride of $2$. The decoder includes 3 spherical convolution layers followed by batch normalization layers, ReLU activation functions, and 3 upsampling layers. The spherical convolution layers have a window size of $3\times3$ and a stride of $1$, while the upsampling layers have a factor of $2$. 
To obtain the output of the \ac{GAN}, the sigmoid activation function has been applied on the last layer. The generator takes the target 360\textdegree\ frame at time $t$ along with the ground-truth saliency map at time $t-k$ to predict the saliency map at time $t$, as done in~\cite{Zhang_ECCV_2018}. The underlying assumption is that, thanks to the availability of the eye-tracker in the headset, the ground truth of the saliency map computed $k$ time instants before can be sent back to the saliency estimator.

The discriminator is composed of 4 standard convolutional layers with kernel size $3\times3$, followed by a fully connected layer. The first 3 layers are paired with ReLU activation functions and have a stride of $2$, while the final layer uses a sigmoid activation function with a stride of $1$. Batch normalization is applied, and dropout with a probability of $0.5$ is introduced to prevent overfitting. The discriminator's role is to distinguish between the predicted and the ground-truth saliency maps at time $t$. The overall network structure is shown in Figure~\ref{fig:GAN}.

\subsection{Loss Function}
The standard GAN loss function, or min-max loss function, is expressed as follows:
\begin{equation}
\begin{split}  
\mathcal{L}_{GAN}(G,D) &= \mathbb{E}_{x}[\log D(x)] + \mathbb{E}_{z}[\log(1 - D(G(z)))]\\ &=\mathbb{E}_{x}[\log D(x)] + \mathbb{E}_{z}[\log(1 - D(\hat{x}))],
\end{split}
\label{cganeq}
\end{equation}
where $z$ is the input of the generator (concatenation of target frame at time $t$ and ground-truth saliency map at time $t-k$), $G(z)$ is the output of the generator (the estimated saliency map $\hat{x}$), and \( D(G(z)) \) is the discriminator’s output for the estimated saliency maps.
In this framework, the generator and discriminator engage in a min-max game: the generator aims to minimize the loss, while the discriminator aims to maximize it. The GAN loss function can be divided into two components: generator loss and discriminator loss, as described in the following.

\subsubsection{Generator Loss}
The generator loss \(\mathcal{L}_{G}\) has been calculated as a combination of four losses: the Pearson linear correlation coefficient loss, \(\mathcal{L}_{\text{CC}}\), the Kullback-Leibler divergence loss, \(\mathcal{L}_{\text{KL}}\), the Spherical \ac{MSE} loss, \(\mathcal{L}_{\text{S\_MSE}}\),
and the \ac{BCE} loss, \(\mathcal{L}_{\text{G\_BCE}}\), as shown in Equation~\ref{eq:1}:

\begin{equation}
\label{eq:1}
\begin{split}
\mathcal{L}_G =& \mathcal{L}_{\text{CC}}(x, \hat{x}) + \mathcal{L}_{\text{KL}}(x, \hat{x}) + \\ &\mathcal{L}_{\text{S\_MSE}}(x, \hat{x}) + \mathcal{L}_{\text{G\_BCE}}(1, \hat{x}), 
\end{split}
\end{equation}
where \( x \) is the ground-truth saliency map.
The CC loss function is defined as \( \mathcal{L}_{\text{CC}}(x,\hat{x}) = 1 - \text{CC}(x,\hat{x}) \), where CC is the Pearson correlation coefficient. Similarly, the KL loss function is defined as \( \mathcal{L}_{\text{KL}}(x,\hat{x}) = \text{KL}(x,\hat{x}) \), where KL is the Kullback–Leibler divergence, and the spherical \ac{MSE} loss is defined as \( L_{\text{S\_MSE}}(x, \hat{x}) = w(\theta, \phi) \times MSE(x, \hat{x}) \), where \(w(\theta, \phi)\) represents spherical weights that incorporate the spatial importance based on spherical geometry by assigning higher weights to regions near the equator and lower weights near the poles. Finally,  
\(\mathcal{L}_{\text{G\_BCE}}\) in the generator quantifies how well the generated data resembles the real data, guiding the generator to produce more realistic outputs. \(\mathcal{L}_{\text{G\_BCE}}\) can be expressed as in Equation~\ref{eq:L_BCE}:
\begin{equation}
\mathcal{L}_{\text{G\_BCE}} = \text{BCE}(\mathbf{1}, D(G(z))),\label{eq:L_BCE}
\end{equation}
where \(\mathbf{1}\) represents the real labels (all ones).

\begin{figure*}[ht!]
     \centering
     \subfloat[]{\includegraphics[width=0.32\linewidth]{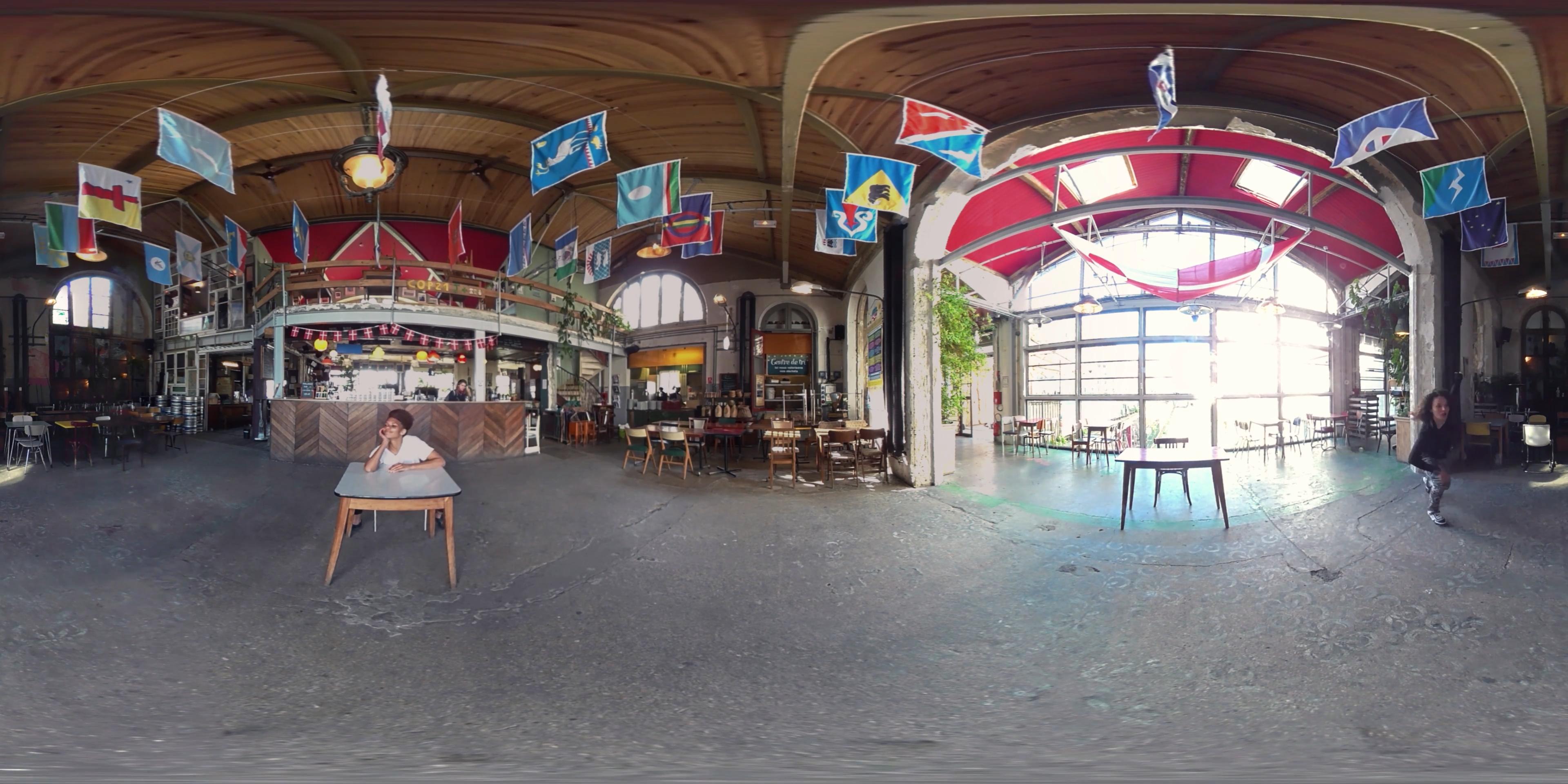}\label{fig:0314}}\hspace{0.2pt}
     \subfloat[]{\includegraphics[width=0.32\linewidth]{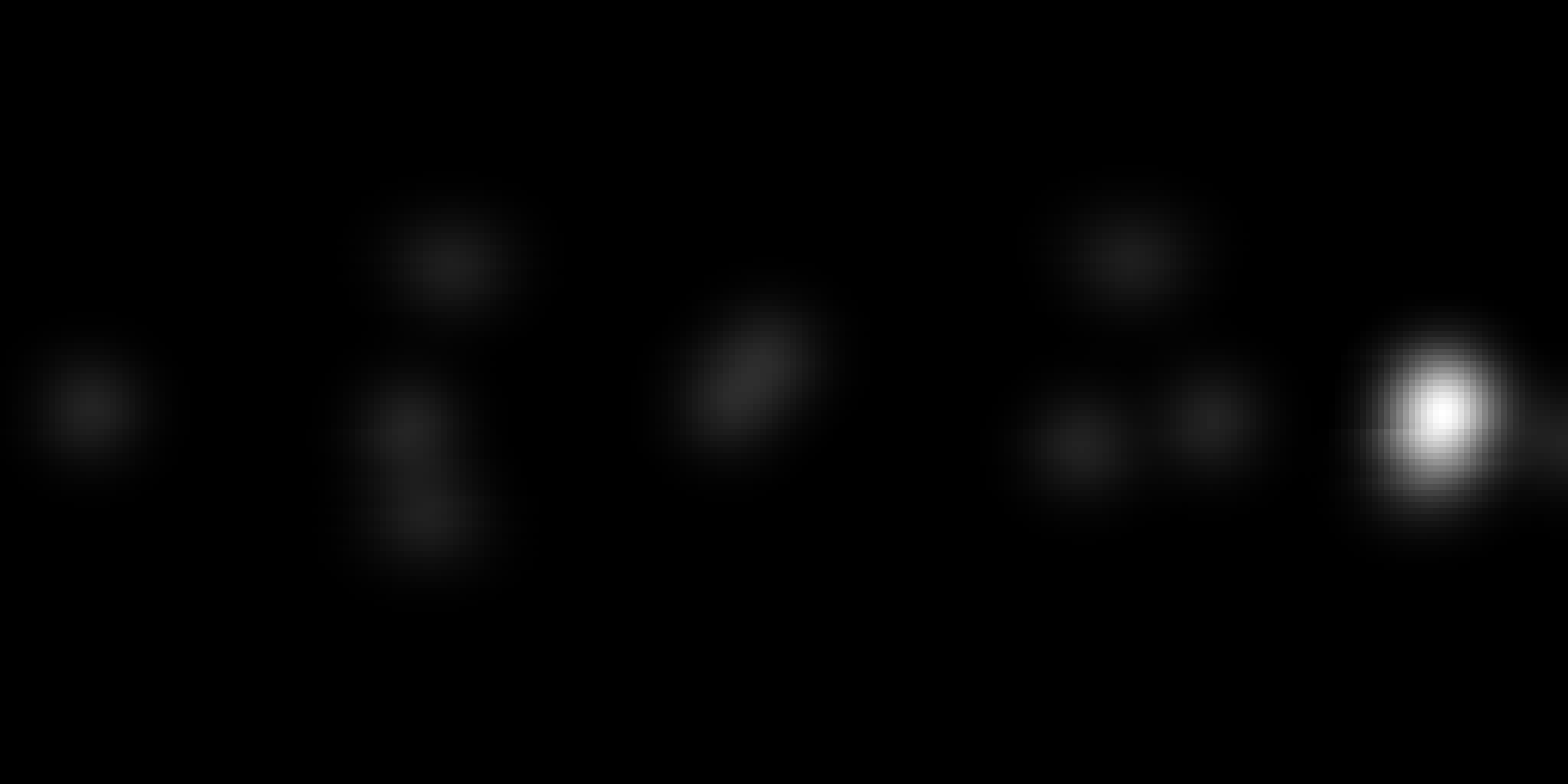}}\label{fig:0314_gt}\hspace{0.2pt}
     \subfloat[]{\includegraphics[width=0.32\linewidth]{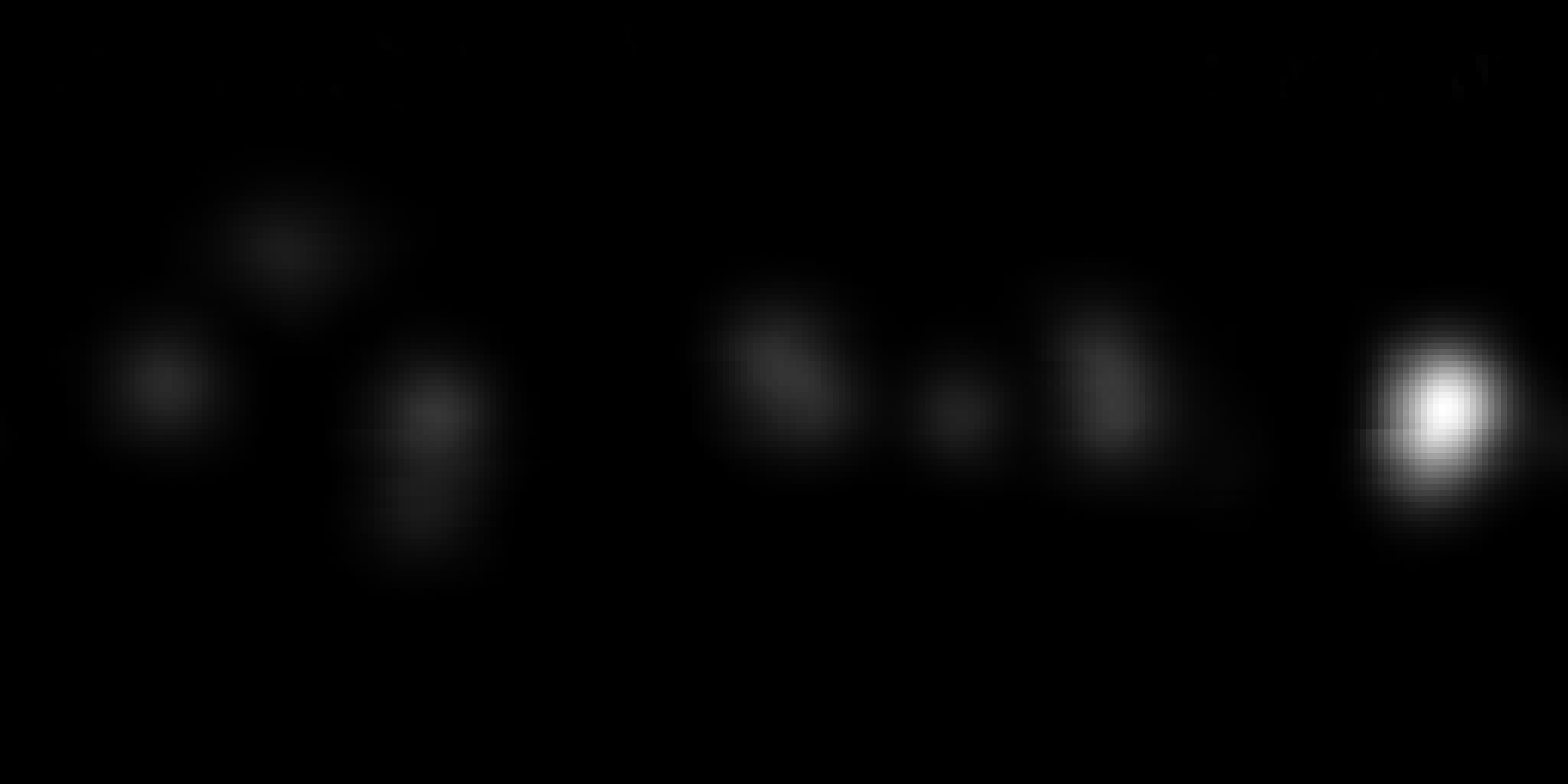}}\label{fig:0314_ours}\\
     \subfloat[]{\includegraphics[width=0.32\linewidth]{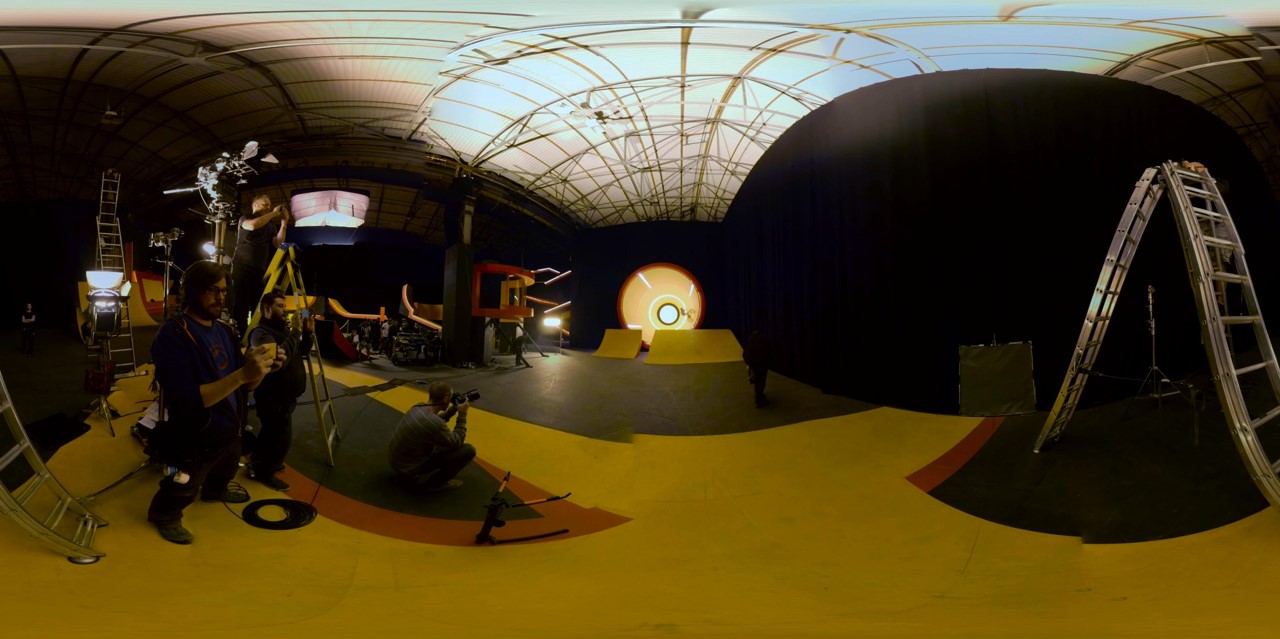}\label{fig:0326}}\hspace{0.2pt}
     \subfloat[]{\includegraphics[width=0.32\linewidth]{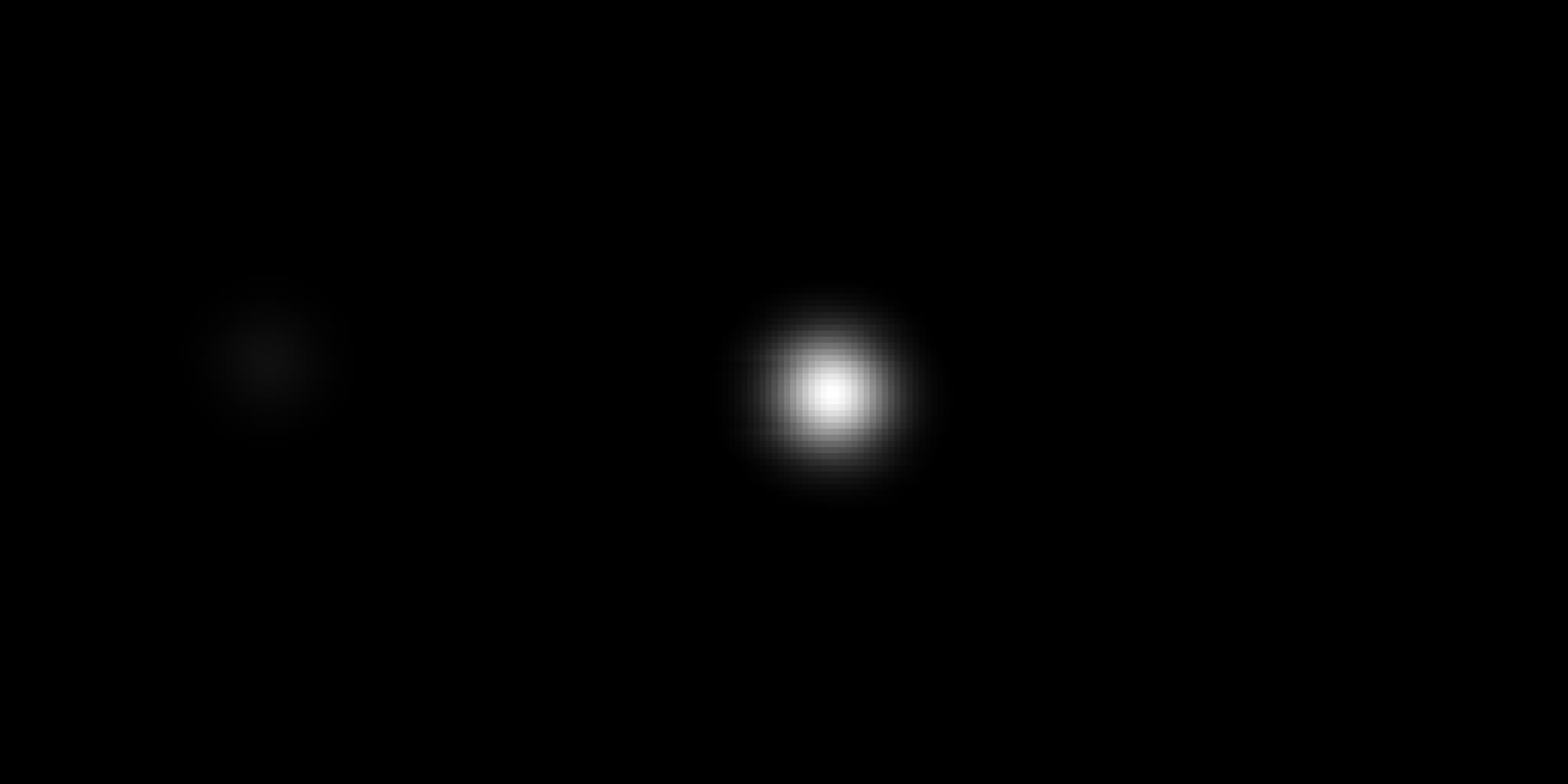}}\label{fig:0326_gt}\hspace{0.2pt}
     \subfloat[]{\includegraphics[width=0.32\linewidth]{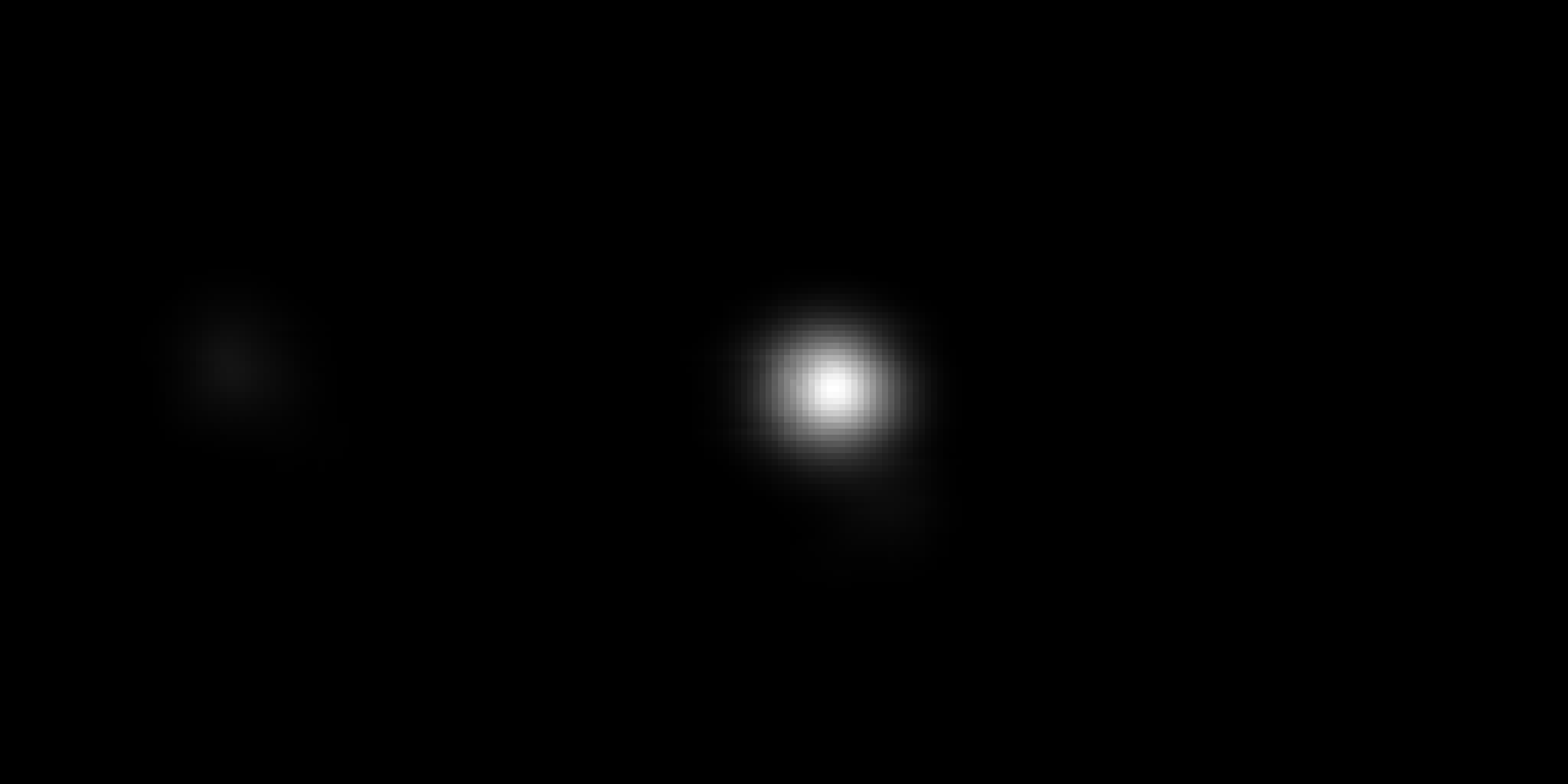}}\label{fig:0326_ours}\\
     \subfloat[]{\includegraphics[width=0.32\linewidth]{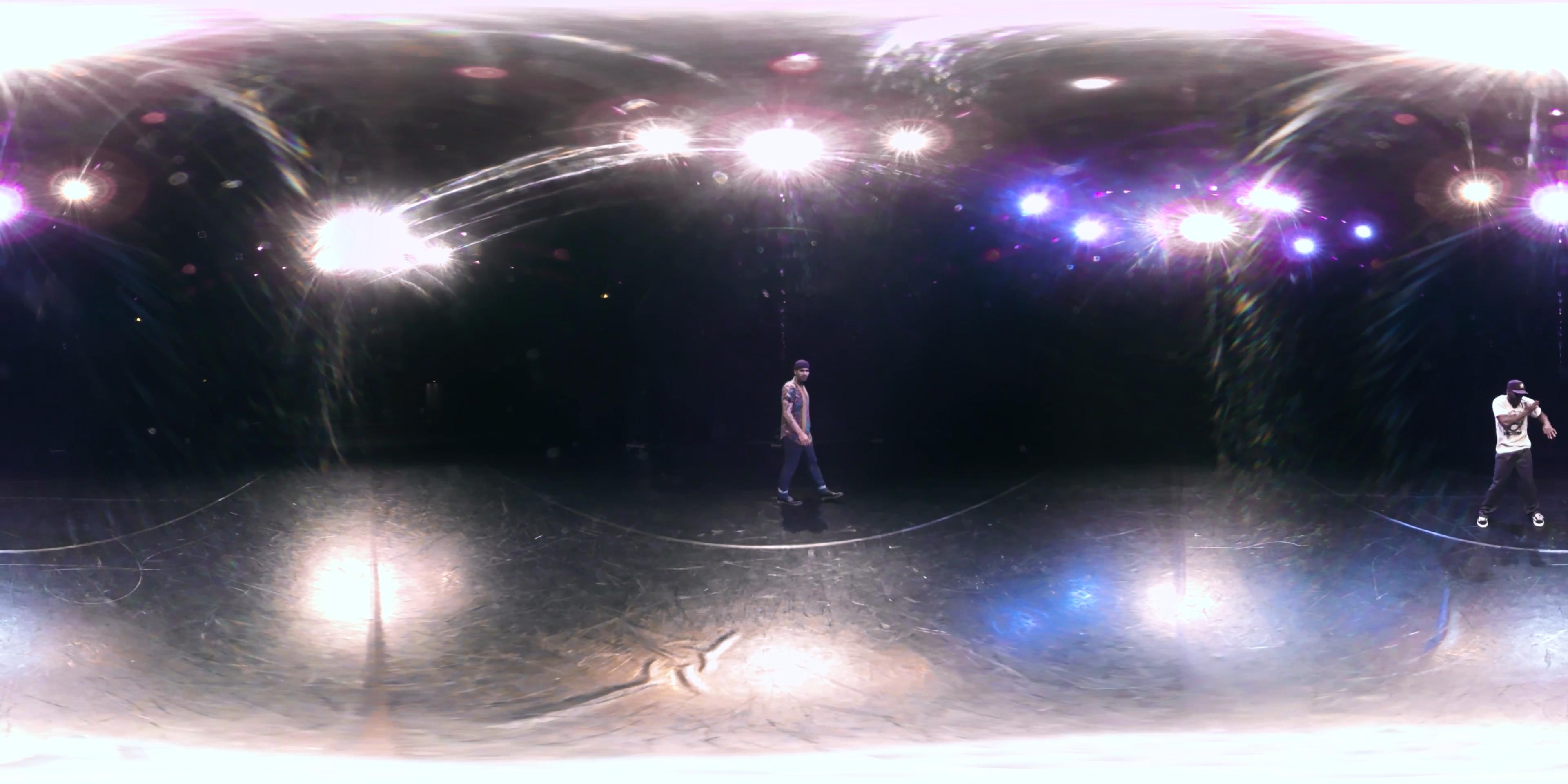}\label{fig:0739}}\hspace{0.2pt}
     \subfloat[]{\includegraphics[width=0.32\linewidth]{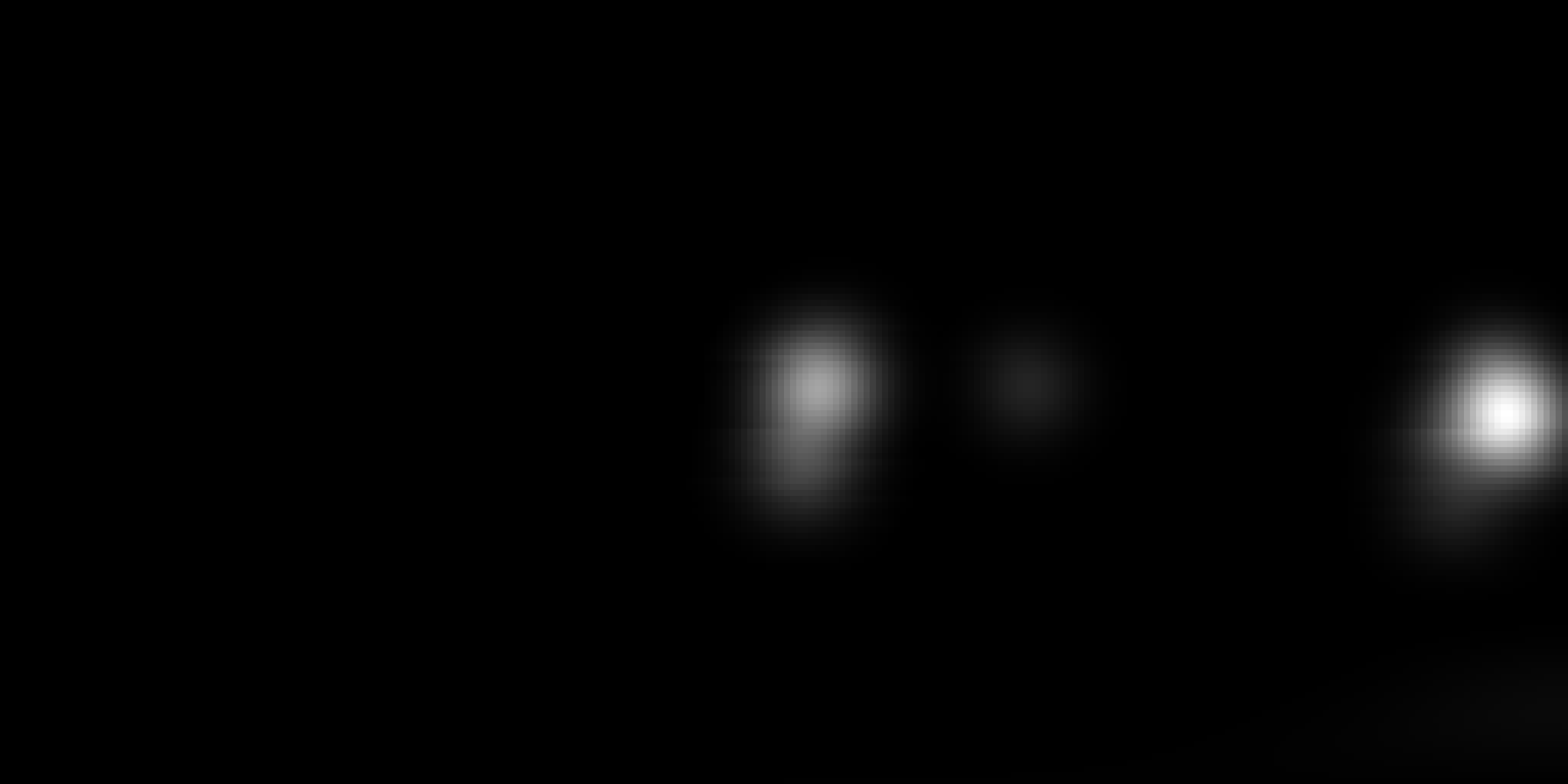}}\label{fig:0739_gt}\hspace{0.2pt}
     \subfloat[]{\includegraphics[width=0.32\linewidth]{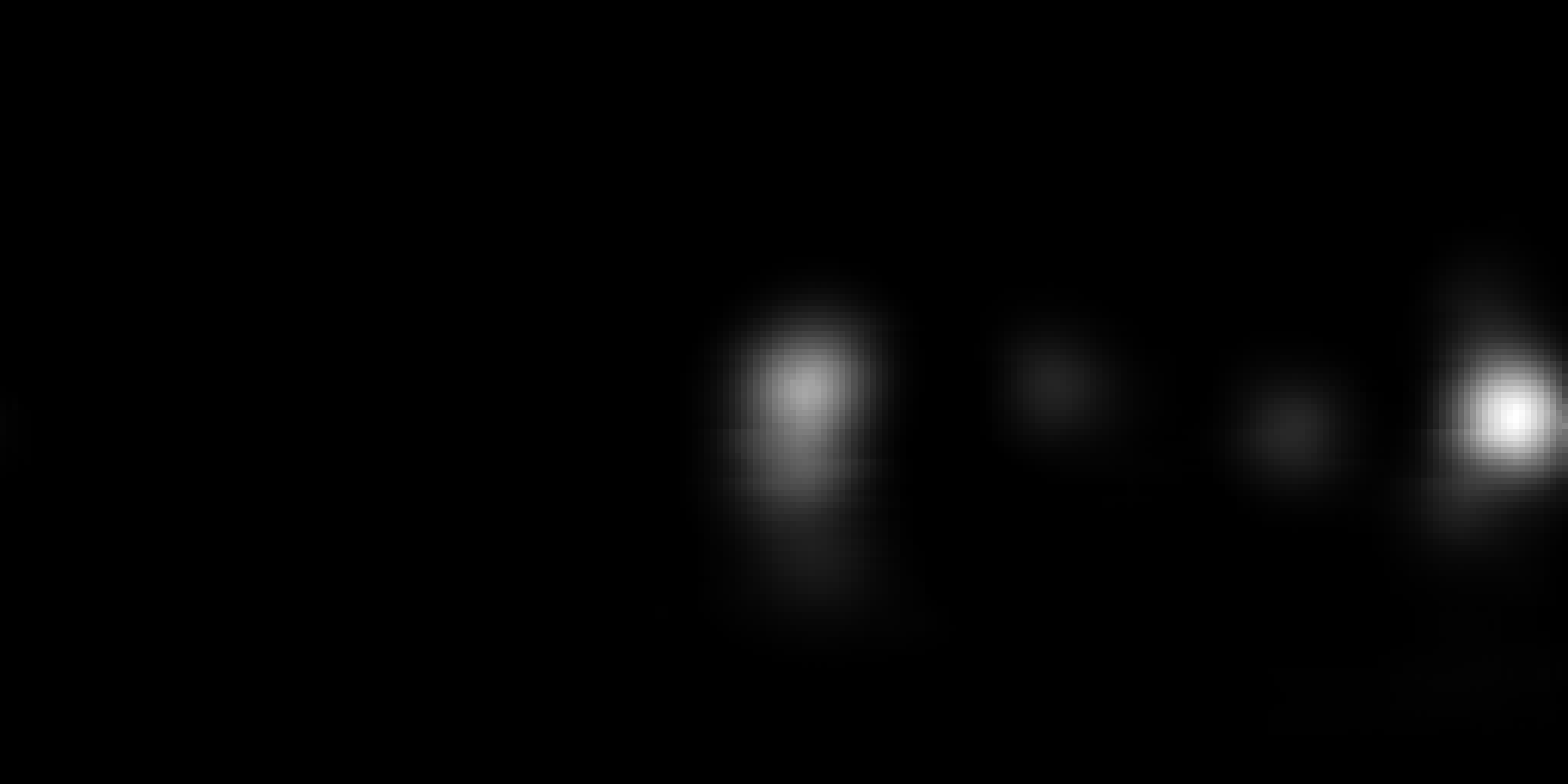}}\label{fig:0739_ours}
     \caption{Qualitative results: (left) three random images extracted from the adopted dataset, (center) ground-truth saliency maps, (right) saliency maps predicted with the proposed method.}
     \label{res}
\end{figure*}

\subsubsection{Discriminator Loss}
To make the discriminator less confident, two strategies have been employed: adding noise to the labels and applying label smoothing. For the first strategy, $10\%$ noise is introduced to both real and fake labels during the calculation of the discriminator loss. The second strategy, label smoothing, is a regularization technique designed to enhance the stability and performance of \ac{GAN} training. This technique involves using slightly perturbed labels instead of hard labels ($0$ and $1$) when training the discriminator. Using label smoothing for real data, the discriminator is encouraged to be slightly less confident about its predictions of real samples, leading to more stable and effective \ac{GAN} training. Label smoothing is applied by multiplying the real labels by $90\%$.

The discriminator loss is calculated as the sum of the \ac{BCE} of the discriminator’s output for the generated saliency map and the \ac{BCE} of the discriminator's output for the ground-truth saliency map. This total loss is then divided by $2$ to reduce the learning rate of the discriminator relative to the generator, as shown in Equation~\ref{eq:3}:
\begin{equation}
\label{eq:3}
\mathcal{L}_{D} = \frac{1}{2} \left( \text{BCE}(\tilde{y}_{\text{real}}, D(x)) + \text{BCE}(\tilde{y}_{\text{fake}}, D(G(z))) \right),
\end{equation}
where:
\begin{align}
\tilde{y}_{\text{real}} &= 0.9 + 0.1 \times \text{rand}(\text{size}(D(x))), \\
\tilde{y}_{\text{fake}} &= 0.1 \times \text{rand}(\text{size}(D(G(z)))).
\end{align}
Specifically, \( \tilde{y}_{\text{real}} \) represents the smoothed and noisy real labels, \( \tilde{y}_{\text{fake}} \) represents the noisy fake labels, 
\( \text{rand}(\text{size}(D(x))) \) generates random values between $0$ and $1$, with the same size as \( D(x) \), and \( \text{rand}(\text{size}(D(G(z)))) \) generates random values between $0$ and $1$, with the same size as \( D(G(z)) \).

\section{Experimental results}
\label{sec:typestyle}

\subsection{Experimental Setup}
For model training, we used He initialization for the weights in both the generator and discriminator networks to ensure stable learning dynamics. In addition, according to the implementation in~\cite{Zhang_ECCV_2018}, we set $k$ equal to $5$. The generator and discriminator are optimized using the Adam optimizer with a learning rate of $1 \times 10^{-4}$ for the generator and $1 \times 10^{-5}$ for the discriminator. The model is trained over $200$ epochs with a batch size of $16$. For testing, we used a batch size of 1.

\subsection{Dataset}
For model training and testing, we used the dataset proposed in~\cite{Zhang_ECCV_2018}. The dataset contains $104$ 360\textdegree\ videos including five sports whose duration varies from $20$ to $60$ seconds. The videos were viewed by $20$ participants using an HTC VIVE equipped with a `7invensun a-Glass' eye tracker. The dataset includes RGB images, ground-truth saliency maps, and gaze points. We used $94$ videos ($68009$ frames) for training and $10$ videos ($8602$ frames) for testing, maintaining an approximate 90\% to 10\% split between the training and testing sets.

\begin{table}[htb]
\captionsetup{justification=centering}
\caption{Performance comparison between Sphere-GAN and baseline models.}
\centering
\adjustbox{max width=\columnwidth}{
\renewcommand*{\arraystretch}{1.2}
\begin{tabular}{|l|c|c|c|c|}
\hline
\multicolumn{1}{|c|}{Methods} & CC ↑            & NSS ↑            & KL ↓            & AUC\_JUDD ↑       \\ \hline
Sphere-GAN   (ours) & \textbf{0.9082} & \textbf{6.6322} & \textbf{0.3896} & \textbf{0.9717} \\ \hline
Standard-GAN (ours) & 0.8856          & 6.0669          & 0.5660          & 0.9571 \\ \hline
SphereU-Net         & 0.8368         & 5.5356          & 2.1261         & 0.8303         \\ \hline
SphericalU-Net      & 0.6246         & 3.5340          & 3.5002         & 0.8977         \\ \hline
PanoSalNet          & 0.4892         & 2.9814          & 13.3442        & 0.6326         \\ \hline
\end{tabular}}
\label{table1}
\end{table}

\subsection{Baseline methods}
For model comparison, we evaluate Sphere-GAN against three baselines, that we selected based on the similarity with the proposed approach:
\begin{itemize}
    \item Our model with standard convolutions, referred to as Standard-GAN in the following.
    \item SphereU-Net~\cite{Peng_IWCMC_2023}: a saliency detection model built on the U-Net framework and spherical \ac{CNN}, focusing on distortion invariance.
    \item SphericalU-Net~\cite{Zhang_ECCV_2018}: a saliency detection model leveraging U-Net and spherical \ac{CNN}, designed to effectively handle variations in spherical image projections, including rotation. 
    \item PanoSalNet~\cite{Nguyen2018}: a nine-layer Deep \ac{CNN} for saliency detection, where the first three layers were initialized with VGGNet~\cite{Simonyan2014} parameters, and transfer learning was applied by replacing the fully connected layers with new layers suitable for the saliency detection task.

\end{itemize}

\begin{table*}[ht!]
\centering
\captionsetup{justification=centering}
\caption{Ablation study for the generator loss.}
\adjustbox{max width=\linewidth}{
\begin{tabular}{|l|c|c|c|c|}
\hline
\multicolumn{1}{|c|}{Loss} & CC↑            & NSS↑            & KL↓            & AUC\_JUDD↑       \\ \hline
$\mathcal{L}_{\text{CC}}(x, \hat{x}) + \mathcal{L}_{\text{G\_BCE}}(x, \hat{x})$ & \textbf{0.9090} & 6.6253 & 0.4433 & 0.9697 \\ \hline
$\mathcal{L}_{KL}(x, \hat{x}) + \mathcal{L}_{\text{G\_BCE}}(x, \hat{x})$         &  0.9042        &  6.6254         &    0.3984     &   0.9704       \\ \hline
$\mathcal{L}_{KL}(x, \hat{x}) + \mathcal{L}_{\text{G\_BCE}}(x, \hat{x}) + \mathcal{L}_{S_{MSE}}(x, \hat{x})$         &  0.9049        &  6.3148         &    0.4115     &   0.9683       \\ \hline
$\mathcal{L}_{\text{CC}}(x, \hat{x}) + \mathcal{L}_{KL}(x, \hat{x}) + \mathcal{L}_{\text{G\_BCE}}(x, \hat{x})$         &  0.9026        &  6.4967         &    0.4597     &   0.9710       \\ \hline
$\mathcal{L}_{\text{CC}}(x, \hat{x}) + \mathcal{L}_{KL}(x, \hat{x}) + \mathcal{L}_{S_{MSE}}(x, \hat{x})+\mathcal{L}_{\text{G\_BCE}}(1, \hat{x})$         &  0.9082        &  \textbf{6.6322}         &    \textbf{0.3896}     &   \textbf{0.9717}       \\ \hline
\end{tabular}}
\label{tab:ablation_study}
\end{table*}

\subsection{Metrics}
To evaluate the predicted saliency maps and compare them with the ground truth and other baselines, we employed four well-established metrics for saliency estimation: Pearson  \ac{CC}, Kullback-Leibler Divergence (KL),  \ac{NSS}, and Area Under the Curve–JUDD (AUC\_JUDD). \ac{CC} and KL were calculated between the predicted saliency maps and the ground-truth saliency maps, while \ac{NSS} and AUC\_JUDD were computed using the predicted saliency maps and ground-truth gaze fixations. The former is a measure of correspondence between saliency maps and ground truth, obtained as
the average normalized saliency at fixation points, while the latter is a variation of the AUC metric specifically defined for saliency estimation~\cite{Bylinskii_PAMI_2019}.

\subsection{Performance Evaluation}
Our Sphere-GAN model was tested on the previously described dataset, and Figure~\ref{res} illustrates three qualitative results for randomly selected samples. 
As can be noticed, our model accurately estimates the saliency maps.
Table~\ref{table1} presents a comparison between the proposed model and baseline methods, with the baseline results sourced from~\cite{Zhang_ECCV_2018}~\cite{Peng_IWCMC_2023}. The results show that our model outperforms the other methods in all evaluation metrics,  providing evidence that the innovation brought by the GAN-based approach represents a key contribution for improving the saliency estimation performance. In addition, the comparison with the \ac{GAN}-based approach without spherical convolutions shows the relevant contribution of this component to handle distortions typical of 360-degree content.


\subsection{Ablation study}

To further highlight the contribution of the \ac{GAN} architecture, we performed an ablation study for the generator loss, comparing the results obtained using different combinations of the constituent losses $\mathcal{L}_{CC}$, $\mathcal{L}_{KL}$, $\mathcal{L}_{S\_MSE}$, and $\mathcal{L}_{G\_BCE}$.
We provide the outcome in Table~\ref{tab:ablation_study}.

The obtained results show that the proposed loss achieves the best performances for all metrics except for correlation, which is slightly higher when only the correlation coefficient loss is added to $\mathcal{L}_{G\_BCE}$. Remarkably, apart from the correlation metric, the use of a distribution-based loss function, such as KL divergence, leads to better results. Moreover, it is interesting to note that the combination of the correlation and the KL divergence loss does not yield better performance. A possible explanation for this phenomenon is that combining CC and KL losses may create conflicting optimization goals, as CC focuses on spatial similarity while KL targets distribution alignment. However, the inclusion of the \ac{MSE} provides stability, balancing these objectives and improving overall results.
This demonstrates that all the components of the proposed loss play a relevant role in training the \ac{GAN}.

\begin{table}[htb]
\captionsetup{justification=centering}
\caption{Performance comparison of Sphere-GAN: estimating saliency based on previous predictions and using previous ground-truth saliency at intervals of N Frames.}
\centering
\adjustbox{max width=\columnwidth}{
\renewcommand*{\arraystretch}{1.1}
\begin{tabular}{|c|c|c|c|c|}
\hline
\multicolumn{1}{|c|}{\# of Frames} & CC ↑            & NSS ↑            & KL ↓            & AUC\_JUDD ↑      \\ \hline
1            & \textbf{0.9082} & \textbf{6.6322} & \textbf{0.3896} & \textbf{0.9717} \\ \hline
2            & 0.8869          & 6.4312          & 6.4312          & 0.9685          \\ \hline
3            & 0.8632         & 6.2071          & 0.5447          & 0.9647         \\ \hline
4           & 0.8377          & 5.9764          & 0.6317          & 0.9605         \\ \hline
5            & 0.8121         & 5.7479          & 0.7191          & 0.9563          \\ \hline
6            &   0.7879       &   5.5397       &   0.8038        &   0.9519        \\ \hline
7            &  0.7631        &     5.3249     &    0.8864       & 0.9477          \\ \hline
8            & 0.7371         & 5.1114          & 0.9741          &  0.9430         \\ \hline
9            & 0.7143         &    4.9208      &     1.0840      &    0.9373       \\ \hline
10            &   0.6914       &    4.7191      &  1.1862         &   0.9321        \\ \hline
\end{tabular}}
\label{table_resN}
\end{table}

Finally, we assessed Sphere-GAN's ability to estimate saliency based on previous predictions instead of using the ground-truth saliency map. Relying solely on the last estimated saliency map throughout the entire video can make accurate prediction difficult due to error accumulation. To address this, we conducted an experiment where the model predicts saliency based on prior estimations, while incorporating the ground-truth saliency map every $N$ frames. More specifically, for $N-1$ frames, the saliency map is estimated based on the saliency computed at time $t-1$. Then, for the $N$-th frame, it is estimated from the ground-truth saliency at $t-5$. In Table~\ref{table_resN}, we present results for different $N$ values ($N = \{1,\,2,\,3,\,...,\,10\}$). When $N = 1$, the model always uses the ground-truth saliency. 
Remarkably, by comparing the results in Table~\ref{table_resN} with the ones provided in Table~\ref{table1}, for all $N$ values, our model outperforms the SpericalU-Net and PanoSalNet. Concerning SphereU-net, SphereGAN always achieves better performances in tems of KL and AUC\_JUDD. In addition, for $N$ in the range $1$-$4$ it achieves also a better CC value, and for $N$ smaller than $7$ it has a higher NSS value.
These findings highlight the model's robustness in predicting saliency maps with reduced reliance on ground-truth data, and the superiority of the \ac{GAN}-based approach.

As a final consideration, we evaluated the complexity of the proposed method with respect to the existing approaches. While our technique introduces complexity compared to simple U-nets, the proposed discriminator only adds $94.72$k parameters. Since SphereU-Net has about $1.7$M parameters according to our calculations, the discriminator only adds about 5.5\% complexity. In addition, it is useful to underline that, in applications like viewport prediction, the model can be deployed on the server side, thus not impacting the edge devices.

\section{Conclusions}
In this work, we presented Sphere-GAN, a novel approach for saliency map prediction of 360\textdegree\ videos. It improves saliency estimation by using a \ac{GAN} architecture with spherical convolutions. Experimental results obtained on a public dataset show that our model outperforms existing methods, demonstrating its potential. 


As future contribution, we plan to develop saliency estimation models that do not rely on ground-truth data. Instead, these models will use only the current and previous 360-degree frames, aiming to generalize saliency prediction across a wide range of scenarios where ground truth may be unavailable.
Furthermore, we intend to incorporate these saliency models into viewport prediction pipelines. This integration is expected to enhance prediction accuracy, reduce bandwidth consumption, and improve user experience, ultimately contributing to more dynamic and immersive omni-directional content delivery.
In addition, the evaluation of the proposed approach will be extended to addition datasets for testing its generalization capabilities.


\bibliographystyle{IEEEbib}
\bibliography{refs}

\end{document}